\documentclass[twoside,11pt]{article}

\usepackage{jmlr2e}
\usepackage{hyperref}

\usepackage{courier}
\usepackage{listings}

\ShortHeadings{Blocks and Fuel}{Van Merri\"{e}nboer et al.}
\firstpageno{1}

\begin{document}

\title{Blocks and Fuel: Frameworks for deep learning}

\author{\name Bart van Merri\"{e}nboer \email bart.van.merrienboer@umontreal.ca \\
        \addr Montreal Institute for Learning Algorithms, University of Montreal, Montreal, Canada
        \AND
        \name Dzmitry Bahdanau \email d.bahdanau@jacobs-university.de \\
        \addr Jacobs University, Bremen, Germany
        \AND
        \name Vincent Dumoulin \email dumouliv@iro.umontreal.ca \\
        \name Dmitriy Serdyuk \email serdyuk@iro.umontreal.ca \\
        \name David Warde-Farley \email wardefar@iro.umontreal.ca \\
        \addr Montreal Institute for Learning Algorithms, University of Montreal, Montreal, Canada
        \AND
        \name Jan Chorowski \email jan.chorowski@ii.uni.wroc.pl \\
        \addr University of Wroc\l aw, Wroc\l aw, Poland
        \AND
        \name Yoshua Bengio \email yoshua.bengio@umontreal.ca \\
        \addr Montreal Institute for Learning Algorithms, University of Montreal, Montreal, Canada \\
        CIFAR Senior Fellow}

% \editor{?}

\maketitle

\begin{abstract}%   <- trailing '%' for backward compatibility of .sty file
  We introduce two Python frameworks to train neural networks on large
  datasets: \emph{Blocks} and \emph{Fuel}. \emph{Blocks} is based on
  Theano, a linear algebra compiler with
  CUDA-support~\citep{Bastien-Theano-2012,bergstra+al:2010-scipy}. It
  facilitates the training of complex neural network models by providing
  parametrized Theano operations, attaching metadata to Theano's symbolic
  computational graph, and providing an extensive set of utilities to assist
  training the networks, e.g.\ training algorithms, logging, monitoring,
  visualization, and serialization. \emph{Fuel} provides a standard format for
  machine learning datasets. It allows the user to easily iterate over large
  datasets, performing many types of pre-processing on the fly.
\end{abstract}


\section{Introduction}

\emph{Blocks} and \emph{Fuel} are being developed by the Montreal Institute of
Learning Algorithms (MILA) at the University of Montreal. Their focus lies on
quick prototyping of complex neural network models. The intended target
audience is researchers who design and experiment machine learning algorithms,
especially deep learning algorithms.

Several other libraries built on top of Theano exist, including Pylearn2
and GroundHog (also developed by MILA), Lasagne, and Keras. Like its
MILA-developed predecessors, Blocks maintains a focus on research and
rapid prototyping. Blocks differentiates itself most notably from the above
mentioned toolkits in its unique relationship with Theano. Instead of
introducing new abstract objects representing `models' or `layers',
Blocks annotates the Theano computational graph, maintaining the flexibility of
Theano while making large models manageable.

Data processing is an integral part of training neural networks, which is not
addressed by many of the aforementioned frameworks. \emph{Fuel} aims to fill
this gap. It provides tools to download datasets and iterate/preprocess them
efficiently.

Both Blocks and Fuel were developed from the very beginning with a strong focus
on software engineering best practices. The development teams strive for
high test coverage, thorough documentation and carefully considered APIs.

\section{Blocks}

Blocks comprises several components, which can be used independently from each
other.

\subsection{Bricks}

Theano is a popular choice for the implementation of neural networks (see
e.g.~\cite{Goodfellow-et-al-ICML2013, Pascanu-et-al-ICML2013}). Blocks and many
other libraries, such as Pylearn2~\citep{pylearn2_arxiv_2013}, build on Theano
by providing reusable components that are common in neural networks, such as
linear transformations followed by non-linear activations, or more complicated
components such as LSTM units. In Blocks these components are referred to as
\emph{bricks} or ``parametrized Theano operations''.

Bricks consist of a set of Theano shared variables, for example the weight
matrix of a linear transformation or the filters of a convolutional layer.
Bricks use these parameters to transform symbolic Theano variables.

Bricks can contain other bricks within them. This introduces a hierarchy on top
of the flat computational graph defined by Theano, which makes it easier to
address and configure complex models programmatically.

The parameters of bricks can be initialized using a variety of schemes that are
popular in the neural network literature, such as sparse initialization,
orthogonal initialization for recurrent weights, etc.

Blocks comes with a large number of `bricks'. Besides standard activations and
transformations used in feedforward networks (maxout, convolutional layers,
table lookups) these also include a variety of more advanced recurrent neural
network components like LSTM, GRU, and support for attention mechanisms (for an
overview of different kinds of network architectures, regularization methods,
and optimization algorithms see~\cite{Bengio-et-al-2015-Book}).

\subsection{Graph management}

Large neural networks can often result in Theano computational graphs containing
hundreds of variables and operations. Blocks does not attempt to abstract away
this complex graph, but to make it manageable by annotating variables in the
graph. Each input, output, and parameter of a brick is annotated as such.
Variables can also be annotated with the role they play in a model, such as
\emph{weights}, \emph{biases}, \emph{filters}, etc.

A series of convenience tools were written that allow users to filter the
symbolic computational graph based on these annotations, and apply transformations
to the graph. Many regularization methods such as weight decay, weight noise, or dropout can be
implemented in a generic, model-agnostic way. Furthermore a complex query mechanism allows 
for their fine-grained application
such as ``apply weight noise to all weights that belong to an LSTM unit whose
parent is a brick with the name \emph{foo}''.

\subsection{Training algorithms}

The gradient descent training algorithm in Blocks is composed of different
`step rules' that modify the descent direction (learning rate scaling,
momentum, gradient clipping, weight norm clipping, etc.). A variety of
algorithms such as AdaGrad, ADADELTA, Adam, RMSProp are available as step
rules.

\subsection{Training}

Experiment management is performed using a `main loop', which combines a Theano
graph with a training algorithm and a Fuel data stream. The main loop has a
flexible extension interface, which is used to perform tasks such as monitoring
on a validation set, serialization, learning rate scheduling, plotting, printing
and saving logs, etc.

\section{Fuel}

Fuel's goal is to provide a common interface to a variety of data formats and
published datasets such as MNIST, CIFAR-10, ImageNet, etc.\ while making it easy
for users to write an interface to new datasets.

Blocks relies on Fuel for its data interface, but Fuel can easily be used by
other machine learning frameworks that interface with datasets.

\subsection{Iteration and preprocessing pipeline}

Fuel allows for different ways of iterating over these datasets, such as
sequential or shuffled minibatches, support for in-memory and out-of-core
datasets, and resampling (cross validation, bootstrapping).

It also provides a variety of on-the-fly preprocessing methods such as random
cropping of images, creating n-grams from text files, and the ability to
implement many other methods easily. These preprocessing steps can be
chained together to form more complex transformations of the input data.

To sidestep Python's global interpreter lock (GIL) and ensure optimal
performance, Fuel can perform all operations in a separate process,
transferring the processed data to the training process using TCP sockets.

\subsection{Standardized data format}

Datasets are distributed in a wide range of formats. Fuel simplifies dataset
storage by converting all built-in datasets to annotated HDF5
files~\citep{hdf5}. In addition to being
an efficient format for large datasets that don't fit into memory, HDF5 is easy
to organize and document. All of the data is stored in a single HDF5 file, with
the following metadata attached:

\begin{itemize}
  \item What are the data sources available (e.g.\ features, targets, etc.)?
  \item How are these data sources officially split (e.g.\ training, validation,
        and test sets)?
  \item Are some data sources unavailable for some splits (e.g.\ test set only
        offers unlabeled examples)?
  \item What are the axes semantics for a given data source (e.g.\ batch,
        feature, width, height, channel, time, etc.)?
\end{itemize}

Integrating user data into Fuel via HDF5 is straightforward, and simply requires
the data to be written to an HDF5 file with metadata according to the specifications.
Finally, while standardizing by convention on HDF5, the Fuel dataset API is
independent of it; users are free to implement dataset objects employing other backends
and use them with the rest of Fuel's components.

\subsection{Automated data management}

Fuel offers built-in scripts that automate the task of downloading datasets,
(similar to e.g.\ \textsc{skdata}\footnote{\url{https://jaberg.github.io/skdata/}})
and converting them to Fuel's HDF5 specification.

The \lstinline$fuel-download$ script is used to download raw data files.
Downloading the raw MNIST data files is as easy as typing
\lstinline$fuel-download mnist$. The \lstinline$fuel-convert$ script is used to
convert raw data files into HDF5-format.

Reproducibility being an important feature of both Fuel and Blocks, the
\lstinline$fuel-convert$ script automatically tags all files it creates with
relevant module and interface versions and the exact command that was used to
generate these files. Inspection of this metadata is done with the
\lstinline$fuel-info$ script.

\section{Serialization and checkpointing}

The training of large, deep neural networks can often take days or even weeks.
Hence, regular checkpointing of training progress is important. Blocks aims to
make the resumption of experiments entirely transparent, even across platforms,
while ensuring the reproducibility of these experiments.

This goal is complicated by shortcomings in Python's \textsc{Pickle}
serialization module, which is unable to serialize many iterators, which Fuel
heavily depends on in order to iterate over large datasets efficiently. To
circumvent this we reimplemented the \textsc{itertools} module from the Python
standard library to be
serializable\footnote{\url{https://github.com/mila-udem/picklable-itertools}}.

As a result, Blocks experiments are able to be interrupted in the middle of a
pass over the dataset, serialized, and resumed later, without affecting the
final training results.

\section{Documentation and community}

Blocks and Fuel are well documented, with both API documentation and tutorials
available online. Two active mailing
lists\footnote{\url{https://groups.google.com/d/forum/blocks-users} and
\url{https://groups.google.com/d/forum/fuel-users}} support users of the
libraries. A separate
repository\footnote{\url{https://github.com/mila-udem/blocks-examples}} is
maintained for users to contribute non-trivial examples of the use of Blocks.
Implementations of neural machine translation models
(NMT,~\cite{bahdanau2015neural}) and the Deep Recurrent Attentive Writer
(DRAW,~\cite{gregor2015draw}) model are publicly available examples of
state-of-the-art models succesfully implemented using Blocks.

\acks{The authors would like to acknowledge the support of the following
agencies for research funding and computing support: NSERC, Calcul Qu\'{e}bec,
Compute Canada, the Canada Research Chairs and CIFAR\@. Bahdanau thanks Planet
Intelligent Systems GmbH for their financial support. We would also like to
thank the developers of Theano.}

\bibliography{bibliography}

\begin{thebibliography}{9}
\providecommand{\natexlab}[1]{#1}
\providecommand{\url}[1]{\texttt{#1}}
\expandafter\ifx\csname urlstyle\endcsname\relax
  \providecommand{\doi}[1]{doi: #1}\else
  \providecommand{\doi}{doi: \begingroup \urlstyle{rm}\Url}\fi

\bibitem[Bahdanau et~al.(2015)Bahdanau, Cho, and Bengio]{bahdanau2015neural}
Dzmitry Bahdanau, Kyunghyun Cho, and Yoshua Bengio.
\newblock Neural machine translation by jointly learning to align and
  translate.
\newblock In \emph{Proceedings of the International Conference on Learning
  Representations (ICLR)}, 2015.

\bibitem[Bastien et~al.(2012)Bastien, Lamblin, Pascanu, Bergstra, Goodfellow,
  Bergeron, Bouchard, and Bengio]{Bastien-Theano-2012}
Fr{\'{e}}d{\'{e}}ric Bastien, Pascal Lamblin, Razvan Pascanu, James Bergstra,
  Ian~J. Goodfellow, Arnaud Bergeron, Nicolas Bouchard, and Yoshua Bengio.
\newblock Theano: new features and speed improvements.
\newblock NIPS Workshop: Deep Learning and Unsupervised Feature Learning, 2012.

\bibitem[Bengio et~al.(2015)Bengio, Goodfellow, and
  Courville]{Bengio-et-al-2015-Book}
Yoshua Bengio, Ian~J. Goodfellow, and Aaron Courville.
\newblock Deep learning.
\newblock Book in preparation for MIT Press, 2015.
\newblock URL \url{http://www.iro.umontreal.ca/~bengioy/dlbook}.

\bibitem[Bergstra et~al.(2010)Bergstra, Breuleux, Bastien, Lamblin, Pascanu,
  Desjardins, Turian, Warde-Farley, and Bengio]{bergstra+al:2010-scipy}
James Bergstra, Olivier Breuleux, Fr{\'{e}}d{\'{e}}ric Bastien, Pascal Lamblin,
  Razvan Pascanu, Guillaume Desjardins, Joseph Turian, David Warde-Farley, and
  Yoshua Bengio.
\newblock Theano: a {CPU} and {GPU} math expression compiler.
\newblock In \emph{Proceedings of the Python for Scientific Computing
  Conference ({SciPy})}, June 2010.

\bibitem[Goodfellow et~al.(2013{\natexlab{a}})Goodfellow, Warde-Farley,
  Lamblin, Dumoulin, Mirza, Pascanu, Bergstra, Bastien, and
  Bengio]{pylearn2_arxiv_2013}
Ian~J. Goodfellow, David Warde-Farley, Pascal Lamblin, Vincent Dumoulin, Mehdi
  Mirza, Razvan Pascanu, James Bergstra, Fr{\'{e}}d{\'{e}}ric Bastien, and
  Yoshua Bengio.
\newblock Pylearn2: a machine learning research library.
\newblock \emph{arXiv preprint arXiv:1308.4214}, 2013{\natexlab{a}}.
\newblock URL \url{http://arxiv.org/abs/1308.4214}.

\bibitem[Goodfellow et~al.(2013{\natexlab{b}})Goodfellow, Warde-Farley, Mirza,
  Courville, and Bengio]{Goodfellow-et-al-ICML2013}
Ian~J. Goodfellow, David Warde-Farley, Mehdi Mirza, Aaron Courville, and Yoshua
  Bengio.
\newblock Maxout networks.
\newblock pages 1319--1327, 2013{\natexlab{b}}.

\bibitem[Gregor et~al.(2015)Gregor, Danihelka, Graves, and
  Wierstra]{gregor2015draw}
Karol Gregor, Ivo Danihelka, Alex Graves, and Daan Wierstra.
\newblock Draw: A recurrent neural network for image generation.
\newblock \emph{arXiv preprint arXiv:1502.04623}, 2015.

\bibitem[Pascanu et~al.(2013)Pascanu, Mikolov, and
  Bengio]{Pascanu-et-al-ICML2013}
Razvan Pascanu, Tomas Mikolov, and Yoshua Bengio.
\newblock On the difficulty of training recurrent neural networks.
\newblock 2013.

\bibitem[{The HDF Group}(1997-2015)]{hdf5}
{The HDF Group}.
\newblock {Hierarchical Data Format, version 5}, 1997-2015.
\newblock http://www.hdfgroup.org/HDF5/.

\end{thebibliography}

\end{document}